\documentclass{article} 
\usepackage{iclr2025_conference,times}

\usepackage{amsmath,amsfonts,bm}

\def\eqref#1{equation~\ref{#1}}

\def\1{\bm{1}}

\def\ra{{\textnormal{a}}}

\def\rx{{\textnormal{x}}}

\def\rva{{\mathbf{a}}}

\def\erva{{\textnormal{a}}}

\def\ervx{{\textnormal{x}}}

\def\rmA{{\mathbf{A}}}

\def\vmu{{\bm{\mu}}}
\def\vtheta{{\bm{\theta}}}
\def\va{{\bm{a}}}

\def\ve{{\bm{e}}}

\def\vx{{\bm{x}}}

\def\eva{{a}}

\def\mA{{\bm{A}}}

\def\mH{{\bm{H}}}
\def\mI{{\bm{I}}}
\def\mJ{{\bm{J}}}

\def\mX{{\bm{X}}}

\def\mSigma{{\bm{\Sigma}}}

\DeclareMathAlphabet{\mathsfit}{\encodingdefault}{\sfdefault}{m}{sl}
\SetMathAlphabet{\mathsfit}{bold}{\encodingdefault}{\sfdefault}{bx}{n}
\newcommand{\tens}[1]{\bm{\mathsfit{#1}}}
\def\tA{{\tens{A}}}

\def\tX{{\tens{X}}}

\def\gG{{\mathcal{G}}}

\def\sA{{\mathbb{A}}}
\def\sB{{\mathbb{B}}}

\def\sS{{\mathbb{S}}}

\def\emA{{A}}

\newcommand{\etens}[1]{\mathsfit{#1}}

\def\etA{{\etens{A}}}

\newcommand{\E}{\mathbb{E}}

\newcommand{\R}{\mathbb{R}}

\newcommand{\KL}{D_{\mathrm{KL}}}
\newcommand{\Var}{\mathrm{Var}}

\newcommand{\Cov}{\mathrm{Cov}}

\newcommand{\normltwo}{L^2}
\newcommand{\normlp}{L^p}

\newcommand{\parents}{Pa} 

\usepackage{booktabs}
\usepackage{hyperref}
\usepackage{url}
\usepackage{algorithm}
\usepackage{algpseudocode}
\usepackage{amsmath}
\usepackage{amssymb}
\usepackage{geometry}
\usepackage{graphicx} 
\usepackage{float}

\title{The Empirical Impact of Reducing Symmetries on the Performance of Deep Ensembles and MoE}

\author{Andrei Chernov \\
Independent Researcher \\
\texttt{chernov.andrey.998@gmail.com} \\
\And
Oleg Novitskij  \\
HSE University \\
\texttt{oanovitskii@edu.hse.ru} \\
}

\iclrfinalcopy 
\begin{document}

\maketitle

\begin{abstract}
Recent studies have shown that reducing symmetries in neural networks enhances linear mode connectivity between networks without requiring parameter space alignment, leading to improved performance in linearly interpolated neural networks. However, in practical applications, neural network interpolation is rarely used; instead, ensembles of networks are more common. In this paper, we empirically investigate the impact of reducing symmetries on the performance of deep ensembles and Mixture of Experts (MoE) across five datasets. Additionally, to explore deeper linear mode connectivity, we introduce the Mixture of Interpolated Experts (MoIE). Our results show that deep ensembles built on asymmetric neural networks achieve significantly better performance as ensemble size increases compared to their symmetric counterparts. In contrast, our experiments do not provide conclusive evidence on whether reducing symmetries affects both MoE and MoIE architectures. The code is available on GitHub\footnote{\url{https://github.com/krds00/asym_ensembles/}}.
\end{abstract}

\section{Introduction}
In the last decade, neural networks have proven to be one of the most important algorithms in the field of machine learning. Despite their undeniable empirical success, many fundamental questions remain unanswered. One such question concerns parameter space symmetries: for any given set of neural network parameters, there exist numerous ‘twins’ that produce exactly the same output for every input while having different parameter values.

There are multiple sources of symmetry in neural network architectures. One prominent example is permutation symmetry in fully connected layers. Consider a standard multi-layer perceptron (MLP). If we swap two neurons in a hidden layer along with their incoming and outgoing weights, the network will produce the exact same output. As a result, any hidden layer of size $n$ has $n!$ different sets of parameters that yield identical outputs. Activation functions such as ReLU \cite{nair2010rectified} can also produce symmetries \cite{wiese2023towards}.

The effects of parameter symmetries have been studied in various areas, including neuron interpretability \cite{godfrey2022symmetries}, optimization \cite{neyshabur2015path}, and Bayesian deep learning \cite{kurle2022detrimental}. In this work, we primarily focus on the impact of symmetries on model accuracy. It has been shown in \cite{lim2024empirical} that eliminating redundant parameters in neural networks improves linear mode connectivity, thereby enhancing the performance of networks whose parameters are obtained by interpolating between two trained models. In this study, we present the Mixture of Interpolated Experts model (see Section \ref{sec:MoIE} for details) and investigate how parameter symmetries influence its performance.

However, the practical application of interpolated neural networks remains controversial, as such architectures usually do not provide a performance boost. The most common approach to leveraging multiple neural networks is through ensembles, where outputs from several models are aggregated to form a final prediction—examples include Deep Ensembles \cite{lakshminarayanan2017simple} and Mixture of Experts (MoE) \cite{fedus2022review}. In this paper, we empirically investigate how reducing symmetry impacts the performance of Deep Ensembles and MoE on $5$ datasets.

\section{Related Work}

\subsection{W-Asymmetric MLP}
Various methods for breaking parameter symmetries in neural networks have been studied, including approaches to removing permutation symmetries \cite{pourzanjani2017improving, pittorino2022deep}, scaling symmetries \cite{badrinarayanan2015understanding}, and sign symmetries \cite{wiese2023towards}. However, in most of these approaches, the neural network architectures or training processes deviate from standard practices, making them difficult to apply in practice. In this work, we fully adopt the approach from \cite{lim2024empirical} to break symmetries in neural networks. This method randomly freezes a portion of the neural network’s weights before training, keeping them unchanged throughout training (see Section \ref{sec:wmlp} for details). Notably, it does not require any special modifications to the training process. Authors of \cite{lim2024empirical} showed that breaking symmetries improves linear mode connectivity between two independently trained neural networks. In this paper, we investigate the empirical impact of reducing symmetries on the performance of Deep Ensembles and Mixture of Experts.

\subsection{Neural Network Ensembles}
In this study, we employ two different approaches for ensembling neural networks. The first approach, known as Deep Ensembles \cite{lakshminarayanan2017simple}, trains $k$ neural networks independently and averages their outputs to obtain the final prediction.

The second approach is the Mixture of Experts (MoE) \cite{yuksel2012twenty}, which consists of two main components: experts and a gating network. Each expert generates an output, but unlike Deep Ensembles, the final prediction is obtained through a weighted average of the experts' outputs. The weights for each expert are dynamically predicted by the gating network rather than being fixed. Recently, MoE architectures utilizing MLP models as experts have gained popularity \cite{fedus2022review} especially in NLP \cite{du2022glam}
and CV domains \cite{puigcerver2023sparse,
riquelme2021scaling}. In this work, we adapt MoE architectures for tabular data from \cite{chernov2025moe}. We cover it in detail in Section \ref{sec:MoE}.

\section{Datasets}

For our work, we selected five datasets to cover different problems:  
\begin{itemize}
    \item \textbf{Regression:} California Housing Prices dataset \cite{pace1997sparse}.
    \item \textbf{Binary classification:} Churn Modeling\footnote{https://www.kaggle.com/shrutimechlearn/churn-modelling} and Adult Income \cite{kohavi1996scaling}.
    \item \textbf{Multi-class classification:} MNIST \cite{deng2012mnist} and Otto Group Product\footnote{https://www.kaggle.com/c/otto-group-product-classification-challenge/data}.
\end{itemize}

Appendix \ref{app:datasets} summarizes the key attributes of these datasets. To ensure consistency, we applied a standardized preprocessing pipeline. Each dataset was split into training, validation, and testing sets with an overall partitioning of 64\% for training, 16\% for validation, and 20\% for testing. Real-valued features were scaled using a \texttt{StandardScaler}, and for classification tasks, the splits were stratified by the target variable. Additional preprocessing steps were applied to each dataset as follows:
\begin{itemize}
    \item \textbf{Churn Modeling dataset:} Non-informative columns such as \texttt{RowNumber}, \texttt{CustomerId}, and \texttt{Surname} were removed.
    \item \textbf{Otto Group Product dataset:} The \texttt{id} column was dropped, and the target variable was encoded using a \texttt{LabelEncoder}.
    \item \textbf{Adult Income dataset:} The target variable was transformed by mapping \texttt{<=\$50K} to $0$ and \texttt{>\$50K} to $1$. Categorical features were processed using a \texttt{OneHotEncoder}, with missing values imputed as \texttt{MissingValue}, while numerical missing values were filled with $0$.
    \item \textbf{California Housing Prices dataset:} Since this dataset contains no missing values, its numerical features were simply scaled.
    \item \textbf{MNIST dataset:} Grayscale images were preprocessed by normalizing and centering pixel values.
\end{itemize}.  

\section{Models}

\subsection{W-Asymmetric MLP}
\label{sec:wmlp}

In this paper, we fully adopt the implementation of W-Asymmetric MLP (WMLP) from \cite{lim2024empirical}, where it was theoretically proven that this approach significantly reduces parameter symmetries. This is achieved by freezing a small portion of the weights, approximately $\mathcal{O}(n^{1/4})$ for details see Algorithm \ref{alg:WMLP}

It is important to emphasize that in ensemble networks utilizing different WMLP models, the frozen neurons—both in value and position—remain identical across all instances. For the hidden layers, we use the GeLU activation function from \cite{hendrycks2016gaussian} in both MLP and WMLP.

\begin{algorithm}[t]
\caption{WMLP Weight and Bias Initialization with Masking}
\label{alg:WMLP}
\begin{algorithmic}[1]
\Require Number of layers \(L = 4\), hidden dimension \(d \in \{64,128,256\}\), and mask seeding parameter \(\mathtt{mask\_num}\).
\State \textbf{Define fixed weights per output unit:}
\[
n_{\text{fix}}^{(1)} = 2,\quad
n_{\text{fix}}^{(l)} =
\begin{cases}
4, & \text{if } d = 256, \\
3, & \text{otherwise}
\end{cases} \quad \text{for } l > 1.
\]
\For{\(l = 1,\dots,L\)}
    \State Let \(W^{(l)} \in \mathbb{R}^{\text{out}_l \times \text{in}_l}\) be the weight matrix.
    \For{\(i = 1,\dots,\text{out}_l\)}
        \State \textbf{Generate Mask:}
        \State For each output unit, select a random subset of \(n_{\text{fix}}^{(l)}\) input indices to  be fixed.The fixed \State positions are determined using a seed based on \(l\) and the output unit index, ensuring \State reproducibility within an ensemble.
        \For{\(j = 1,\dots,\text{in}_l\)}
            \If{\(j\) is in the fixed subset for unit \(i\)}
                \State Set \(W^{(l)}_{ij} \sim \mathcal{N}(0,1)\). The random seed for fixed weights depends on the layer \(l\) and \State weight position. These weights are then frozen.
            \Else
                \State Initialize \(W^{(l)}_{ij}\) using Kaiming Uniform Initialization with parameter \(\sqrt{5}\). The random \State seed for non-frozen weights depends on the repetition number, the estimator index in the \State ensemble, \(l\), and the weight's position.
            \EndIf
        \EndFor
    \EndFor
    \State \textbf{Initialize Bias:} \\
    \hspace*{1em} Set \(b^{(l)}\) uniformly in 
    \[
      \left[-\frac{1}{\sqrt{\text{in}_l}},\ \frac{1}{\sqrt{\text{in}_l}}\right],
    \]
    \Comment{The random seed for bias initialization depends on the repetition number, the estimator index, \(l\), and the bias position.}
\EndFor
\end{algorithmic}
\end{algorithm}

\subsection{Mixture of Experts}
\label{sec:MoE}

In \cite{chernov2025moe}, it was shown that MoE performs at least as well as a vanilla MLP on tabular data while requiring significantly fewer parameters. In this paper, we compare the performance of MoE with MLP as experts against MoE with WMLP experts. 

From \cite{chernov2025moe}, we utilize both the vanilla MoE, where logistic regression is used as a gating neural network, and the Gumbel Gating MoE (GG MoE), which employs the Gumbel-softmax function instead of the standard Softmax activation for logistic regression. Following the original paper, we use $10$ samples from the Gumbel-softmax distribution during inference.

\subsection{Mixture of Interpolated Experts}
\label{sec:MoIE}

Since \cite{lim2024empirical} demonstrated that reducing symmetries improves the performance of linearly interpolated neural networks, we evaluate the performance of the Mixture of Interpolated Experts (MOIE). MOIE uses the same gating function as MoE but, instead of computing a weighted average of the final outputs, it linearly interpolates the weights of the experts to produce an output:

$$\hat{y} = \text{Expert architecture} \left( \sum_{i}^{k} \alpha_i(x) W_i(x) \right),$$

where $\hat{y}$ is the final prediction, $k$ is the number of experts, $\alpha$ is an output from a gating network, and $W_i$ represents the model parameters of each expert. The expert architecture is selected from ${\text{MLP}, \text{WMLP}}$.

\section{Experiments}
\label{Experiments}

\subsection{Setup}
In this section, we describe the details of the training and evaluation procedures applied to Deep Ensembles (Section \ref{Deep ensemble}), MoE and MoIE (Section \ref{MOE MOIE}).

\subsubsection{Deep Ensemble}
\label{Deep ensemble}
We trained models with a batch size of $256$. For constructing Deep Ensembles, we trained $64$ instances of both the MLP and WMLP models, each initialized with a different random seed to ensure variability in the free weights. 

For WMLP, a fixed number of random weights per row, denoted as \(n_{\text{fix}}\), was selected and frozen in each layer. These frozen weights were sampled from a \(\mathcal{N}(0,I)\) distribution. To reduce variance in the final metrics, we repeated training and evaluation $10$ times independently and reported the average evaluation metrics on the test sets.

For Deep Ensembles, we utilized MLP and WMLP blocks. Their structure consisted of an input layer that mapped the number of dataset features to a hidden dimension (\textit{hidden\_dim}), followed by two hidden layers of size \textit{hidden\_dim} \(\times\) \textit{hidden\_dim}, and an output layer of size \textit{hidden\_dim} \(\times\) \textit{out\_features}, where \textit{out\_features} was set to $1$ for regression and to the number of classes for classification. Experiments for Deep Ensembles were conducted for \textit{hidden\_dim} values of $64$, $128$, and $256$.

Loss functions were selected based on the task: \texttt{MSELoss} for regression and \texttt{CrossEntropyLoss} for classification, with \texttt{RMSE} and \texttt{accuracy} serving as the evaluation metrics, respectively. Optimization was carried out using the AdamW optimizer with a learning rate of \(1\times10^{-3}\) and a weight decay of \(3\times10^{-2}\). Each network was trained for up to $1000$ epochs, with $batch\_size=256$ and early stopping triggered if the validation loss did not improve for $16$ consecutive epochs. Training was performed in parallel on 64 CPUs. After each training iteration, we logged the training time, the number of epochs executed, and the performance metric for each of the $64$ MLP and WMLP models. Finally, the individual models were aggregated into Deep Ensembles of $2$, $4$, $8$, $16$, $32$, and $64$ networks. For regression tasks, ensemble predictions were computed as the mean of the individual outputs, while for classification tasks the logits were averaged and the final prediction was determined via the \texttt{argmax} function. For each ensemble, both the ensemble performance metric and an interpolation metric—derived from averaging the model weights—were recorded.

\subsubsection{MoE and MOIE}
\label{MOE MOIE}
In experiments with MoE and MoIE, we used both MLP and WMLP architectures, along with the same loss functions, evaluation metrics, training procedures, and optimizer parameters as described in Section \ref{Deep ensemble}. For these experiments, the expert hidden dimension was fixed at $64$. In the WMLP architecture, the number of fixed weights per output unit, \(n_{\text{fix}}\), was set to $2$ for the input layer and $3$ for subsequent layers. The number of experts was varied among \([2,\,4,\,8,\,16,\,32,\,64]\). We conducted experiments for all models described in Sections \ref{sec:MoE} and \ref{sec:MoIE}.

\subsection{Results}
\label{results}

\begin{figure}[H]
  \begin{center}
    \includegraphics[width=\linewidth, height=0.8\textheight, keepaspectratio]{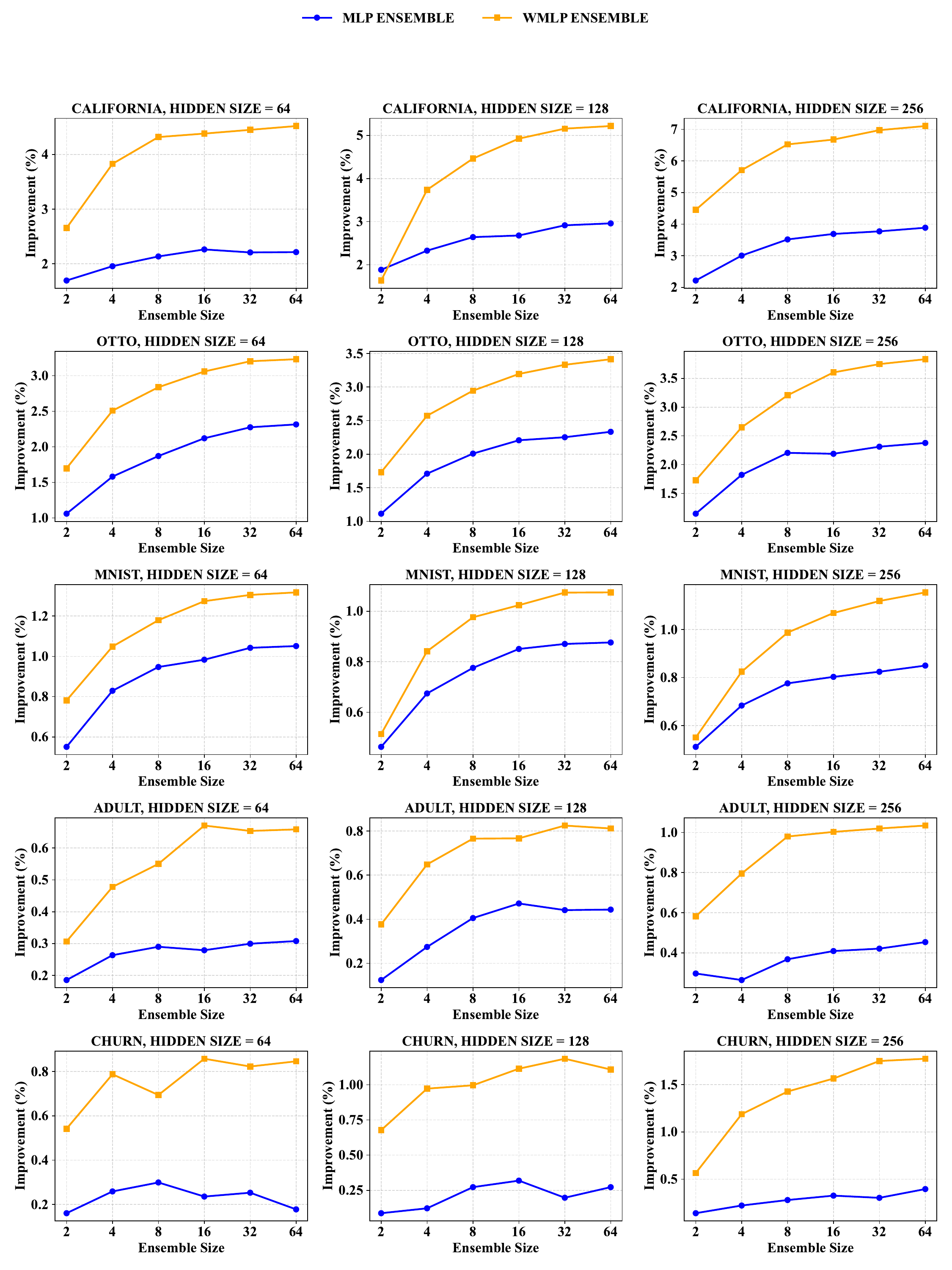}
  \end{center}
  \vspace{1em}
  \caption{Deep ensembles’ relative improvement in performance. The graphics depicts the relative improvement in performance of both MLP and WMLP models compared to a single MLP and WMLP neural network, respectively.}
  \label{fig:deep_relative}
\end{figure}

Figure \ref{fig:deep_relative} presents the experimental results, showing the average performance improvements in test metrics across different random seeds. Specifically, we report accuracy for classification tasks and RMSE for regression tasks, measuring improvement relative to the average performance of a single neural network. The results are presented for each dataset and hidden dimension configuration and indicate that Deep Ensembles with WMLP models improve significantly more than with MLP models and this improvement increases as the ensemble size increases.

A possible explanation for this behavior could be that WMLP deep ensembles perform worse than MLP ensembles in terms of absolute test metric values. However, this is not the case, as demonstrated in Appendix \ref{app:performance}. Given that WMLP models retain the universal approximation property, as shown in \cite{lim2024empirical}, we believe this is a promising finding that could encourage the adoption of asymmetric neural networks in ensembles for practical applications.  

\begin{figure}[H]
  \begin{center}
    \includegraphics[width=\linewidth, height=0.8\textheight, keepaspectratio]{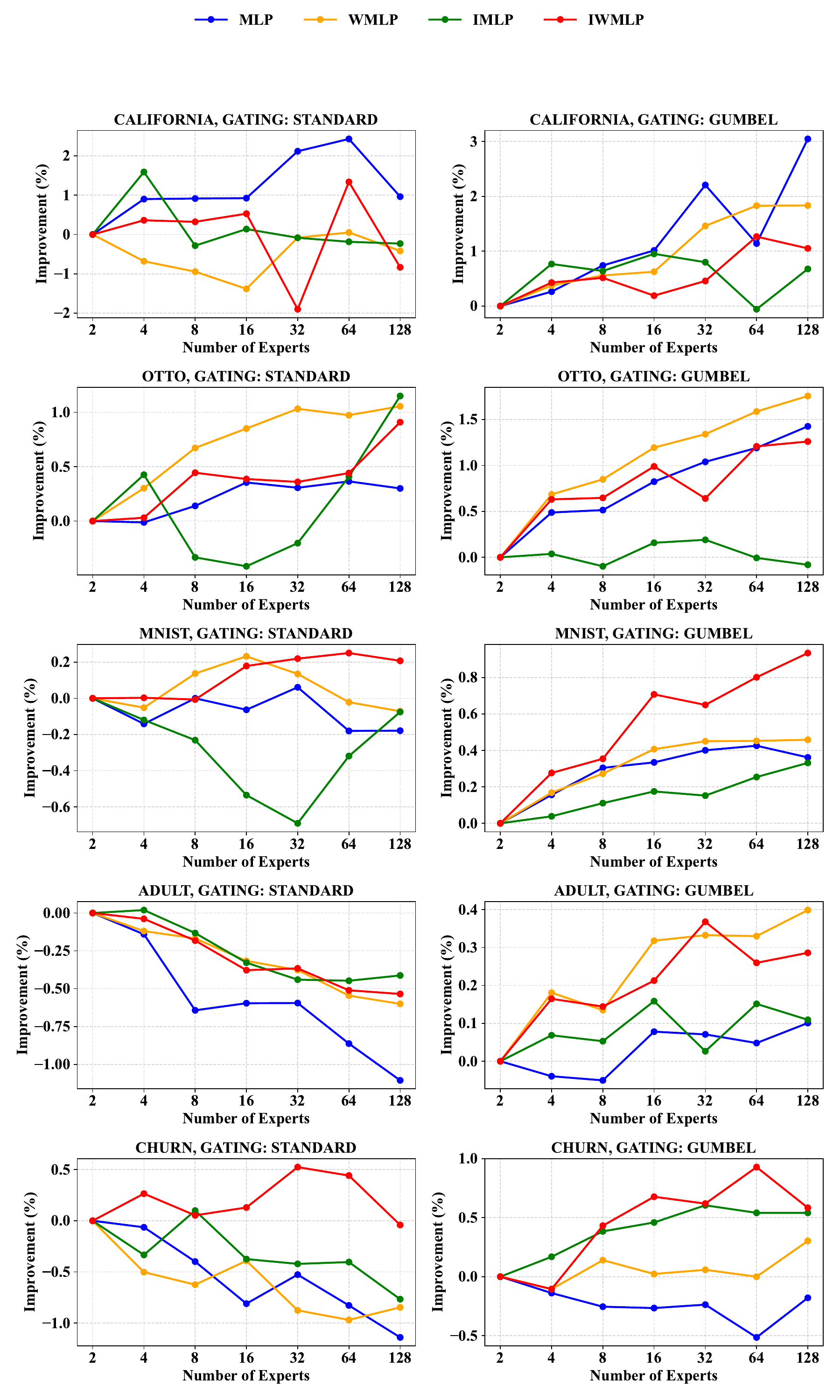}
  \end{center}
  \vspace{1em}
  \caption{MoE and MoIE relative improvement. In these graphics, MLP represents MoE with vanilla MLP experts, WMLP denotes MoE with WMLP experts, IMLP corresponds to MoIE with vanilla MLP experts, and IWMLP refers to MoIE with WMLP experts. The relative improvement of all models is shown in comparison to their corresponding model architectures with two experts.}
  \label{fig:moe_relative}
\end{figure}

Likewise, Figure \ref{fig:moe_relative} shows the relative performance of MoE and MoIE with a varying number of experts compared to their corresponding models with two experts. As discussed in Section \ref{MOE MOIE}, the hidden size of each expert remains constant. Although MoE with WMLP experts and MoIE with WMLP experts tend to outperform their counterparts with MLP experts on $4$ out of $5$ datasets, the improvements are less convincing compared to Deep Ensembles. We also report absolute metrics in Appendix \ref{app:performance}.

One potential reason for unclear results in MoE and MoIE might be that the models tend to overfit, meaning that as the number of parameters increases, test metrics deteriorate. We did not apply any regularization techniques to the neural network architectures to avoid overcomplicating the analysis. However, addressing this issue is essential, and the experimental setup for MoE and MoIE should be adjusted in future work.

\section{Conclusion}
In this paper, we empirically demonstrated that the performance of Deep Ensembles improves significantly with increasing ensemble size when using W-Asymmetric MLP models compared to vanilla MLP models. This result may serve as a first step toward understanding the practical impact of reducing symmetry in neural networks.

However, based on our experiments, we cannot conclude that W-Asymmetric MLP improves the performance of either the Mixture of Experts (MoE) or the Mixture of Interpolated Experts (MoIE) models. As discussed in Section \ref{results}, the experimental setup for MoE should be refined in future work
\newpage

\bibliography{./iclr2025_conference}
\bibliographystyle{iclr2025_conference}

\appendix
\section{Appendix}
\subsection{Summary of Datasets}
\label{app:datasets}
Table \ref{table:datasets} provides a summary of the datasets used in this paper.

\begin{table}[ht]
\centering
\caption{Datasets description}
\label{table:datasets}
\resizebox{\textwidth}{!}{%
\begin{tabular}{@{}p{2.5cm}p{2.5cm}p{2cm}p{3.5cm}p{2.5cm}@{}}
\toprule
\textbf{Dataset} & \textbf{Task} & \textbf{Instances} & \textbf{Feature Details} & \textbf{Target Variable}  \\ \midrule
Churn Modelling & Binary Classification & 10,000 & Customer attributes (e.g., CreditScore, flagGeography, Gender, Age, Tenure, Balance, NumOfProducts) & Customer churn (Exited: yes/no) \\[1ex]
Otto Group Product & Multi-class Classification & 61,878 & 93 anonymized numerical features (feat\_1 to feat\_93) & Product category (9 classes) \\[1ex]
Adult Income & Binary Classification & 48,842 & Mixed continuous and categorical variables (e.g., age, workclass, education, occupation, etc.) & Income level (``\texttt{>50K}'' as 1, ``\texttt{<=50K}'' as 0) \\[1ex]
California Housing Prices & Regression & 20,640 & 8 numerical predictors (MedInc, HouseAge, AveRooms, AveBedrms, Population, AveOccup, Latitude, Longitude) & Median house value (in \$100K units) \\[1ex]
MNIST & Multi-class Classification & 70,000 & Preprocessed rectified into a single vector 28$\times$28 pixel grayscale images (784 features per image) & Handwritten digit (10 classes: 0--9) \\ \bottomrule
\end{tabular}
}
\end{table}

\subsection{Absolute Metrics for all Models on Test Dataset}
\label{app:performance}

\begin{figure}[H]
  \begin{center}
    \includegraphics[width=\linewidth]{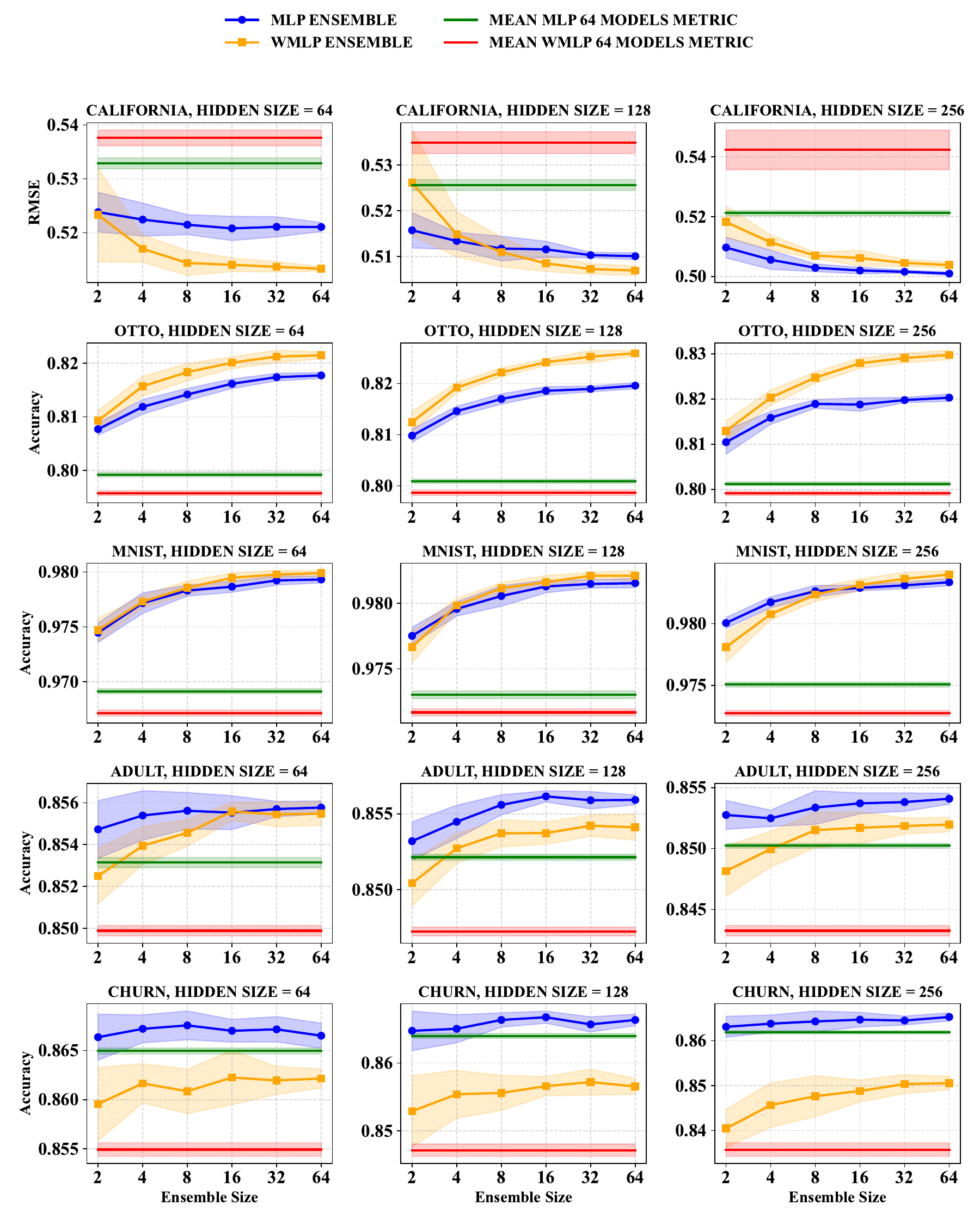}
  \end{center}
  \vspace{1em}
  \caption{Deep ensemble absolute metrics.}
  \label{fig:deep_absolute}
\end{figure}

\begin{figure}[H]
  \begin{center}
    \includegraphics[width=\linewidth]{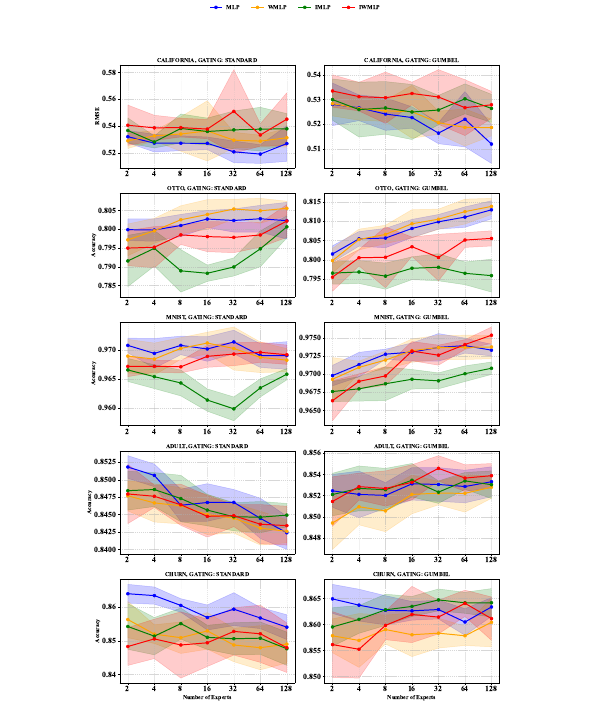}
  \end{center}
  \vspace{1em}
  \caption{MoE/MoIE absolute metrics.}
  \label{fig:moe_absolute}
\end{figure}

Figures~\ref{fig:deep_absolute} and~\ref{fig:moe_absolute} present the absolute results of the experiments described in Section~\ref{Experiments}. In Figure~\ref{fig:deep_absolute}, the change of the relevant metric for deep ensembles using both MLP and WMLP models is shown as a function of ensemble size. The bold lines indicate the mean performance across different random seeds, while the shaded regions represent the $\pm$ one standard deviation intervals. Additionally, the figure displays the mean metric values and intervals for a baseline, which were calculated by aggregating the test metrics of $64$ single MLP and $64$ single WMLP models; these baseline values were subsequently used in Figure~\ref{fig:deep_relative}. It can be observed that the metric improves as the ensemble size increases. Notably, although WMLP models may yield inferior performance when used individually, the WMLP ensemble tends to outperform the MLP ensemble.

Similarly, Figure~\ref{fig:moe_absolute} illustrates the change of the corresponding metric for MoE and MoIE models as a function of the number of experts. Analogous to Figure~\ref{fig:deep_absolute}, the plot shows the mean performance along with the $\pm$ standard deviation intervals obtained from aggregating results over various random seeds. The baseline values corresponding to the case of two experts were used to compute the relative improvements shown in Figure~\ref{fig:moe_relative}. It is evident that the use of Gumbel-softmax leads to better performance compared to the standard softmax, and, in most cases, MoE with WMLP experts or MoIE with WMLP (IWMLP) achieves higher quality than MoE with MLP experts or MoIE with MLP/WMLP, respectively.

\end{document}